\documentclass[conference]{IEEEtran}
\IEEEoverridecommandlockouts
\usepackage{cite}
\usepackage{amsmath,amssymb,amsfonts}
\usepackage{algorithmic}
\usepackage{graphicx}
\usepackage{textcomp}
\usepackage{xcolor}
\def\BibTeX{{\rm B\kern-.05em{\sc i\kern-.025em b}\kern-.08em
    T\kern-.1667em\lower.7ex\hbox{E}\kern-.125emX}}
\begin{document}

\title{Devising a solution to the problems of Cancer awareness in Telangana\\

}
\makeatletter
\newcommand{\linebreakand}{%
  \end{@IEEEauthorhalign}
  \hfill\mbox{}\par
  \mbox{}\hfill\begin{@IEEEauthorhalign}
}
\makeatother
\author{\IEEEauthorblockN{Priyanka Avhad}
\IEEEauthorblockA{\textit{Department of Computer Engg.} \\
\textit{Veermata Jijabai Technological Institute}\\
Mumbai, India 400019 \\
pravhad\_b19@ce.vjti.ac.in }
\and
\IEEEauthorblockN{Vedanti Kshirsagar}
\IEEEauthorblockA{\textit{Department of Computer Engg.} \\
\textit{Veermata Jijabai Technological Institute}\\
Mumbai, India 400019 \\
vakshirsagar\_b19@ce.vjti.ac.in }
\and
\IEEEauthorblockN{Mahek Nakhua}
\IEEEauthorblockA{\textit{Department of Computer Engg.} \\
\textit{Veermata Jijabai Technological Institute}\\
Mumbai, India 400019 \\
msnakhua\_b19@ce.vjti.ac.in }
\linebreakand
\IEEEauthorblockN{Urvi Ranjan}
\IEEEauthorblockA{\textit{Department of Electronics Engg.} \\
\textit{Veermata Jijabai Technological Institute}\\
Mumbai, India 400019 \\
urranjan\_b19@el.vjti.ac.in }
}

\maketitle

\begin{abstract}
According to the data[1], the percent of women who underwent screening for cervical cancer, breast and oral cancer in Telangana in the year 2019-2020 was 3.3 percent, 0.3 percent and 2.3 percent respectively. Although early detection is the only way to reduce morbidity and mortality, people have very low awareness about cervical and breast cancer signs and symptoms and screening practices. We developed an ML classification model to predict if a person is susceptible to breast or cervical cancer based on demographic factors. We devised a system to provide suggestions for the nearest hospital or Cancer treatment centres based on the user's location or address. In addition to this, we can integrate the health card to maintain medical records of all individuals and conduct awareness drives and campaigns. For ML classification models, we used decision tree classification and support vector classification algorithms for cervical cancer susceptibility and breast cancer susceptibility respectively. Thus, by devising this solution we come one step closer to our goal which is spreading cancer awareness, thereby, decreasing the cancer mortality and increasing cancer literacy among the people of Telangana.
\end{abstract}

\section{Introduction}
Cancer Literacy is essential to reduce global cancer mortality and crucial for the detection of cancer at an early stage. Breast Cancer is the most common type of cancer women are diagnosed with worldwide, including in India where advanced stages of diagnosis and rising incidence and mortality rates make it imperative to increase cancer literacy in women.

The NFHS dataset[1] was examined for the indicators, Screening for Cancer among Women (age 30-49 years) in case of cervical cancer, breast cancer, oral cancer. This was classified into two categories - Residence in rural and urban areas, District-wise.

The percentages of screening for Cancer among Women (age 30-49 years) in rural and urban areas are as shown in Fig.1

\begin{figure}[!h]
\centering
\includegraphics[width=3.4in]{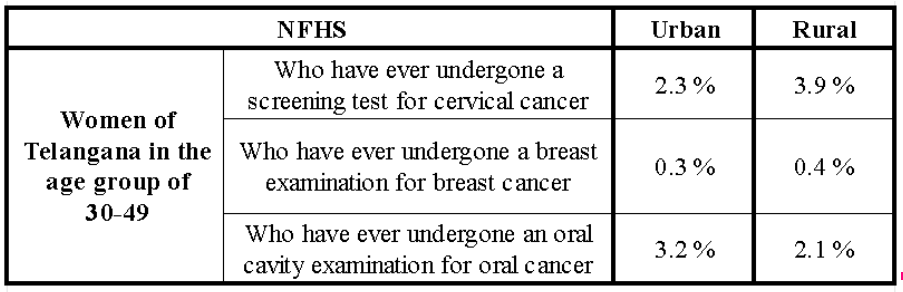}
\caption{\centering{Summary of NFHS dataset highlighting the small percentage of women undergoing these tests}}
\end{figure}

To make a comparison based on districts in Telangana, a new combined dataset was designed with all latitudes and longitudes. Data visualization of 3 indicators (Ever undergone a screening test for cervical cancer (\%),  Ever undergone a breast examination for breast cancer (\%), Ever undergone an oral cavity examination for oral cancer (\%)) was performed. Fig. 2, Fig. 3, and Fig.4 shows results mapped onto the district along with a tabular representation of all the districts in decreasing order of percentages.

\begin{figure}[!h]
\centering
\includegraphics[width=3.4in]{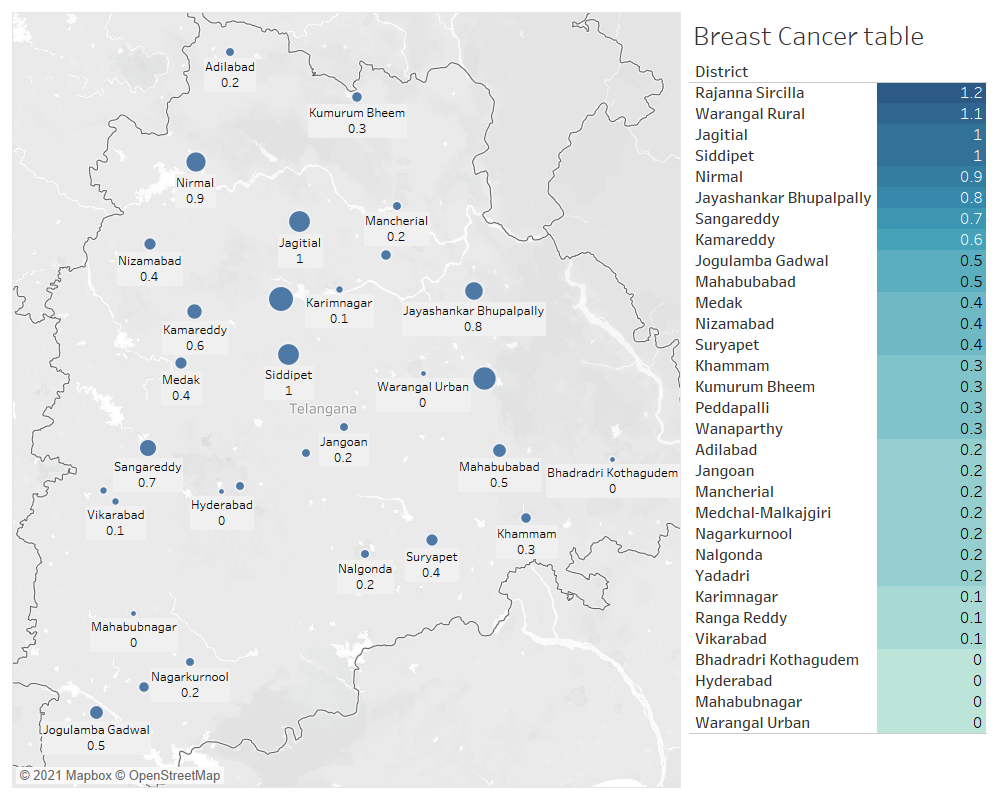}
\caption{\centering{Percentage distribution of Breast Cancer examination tests taken by Women in Telangana (District wise)}}
\end{figure}

\begin{figure}[!h]
\centering
\includegraphics[width=3.4in]{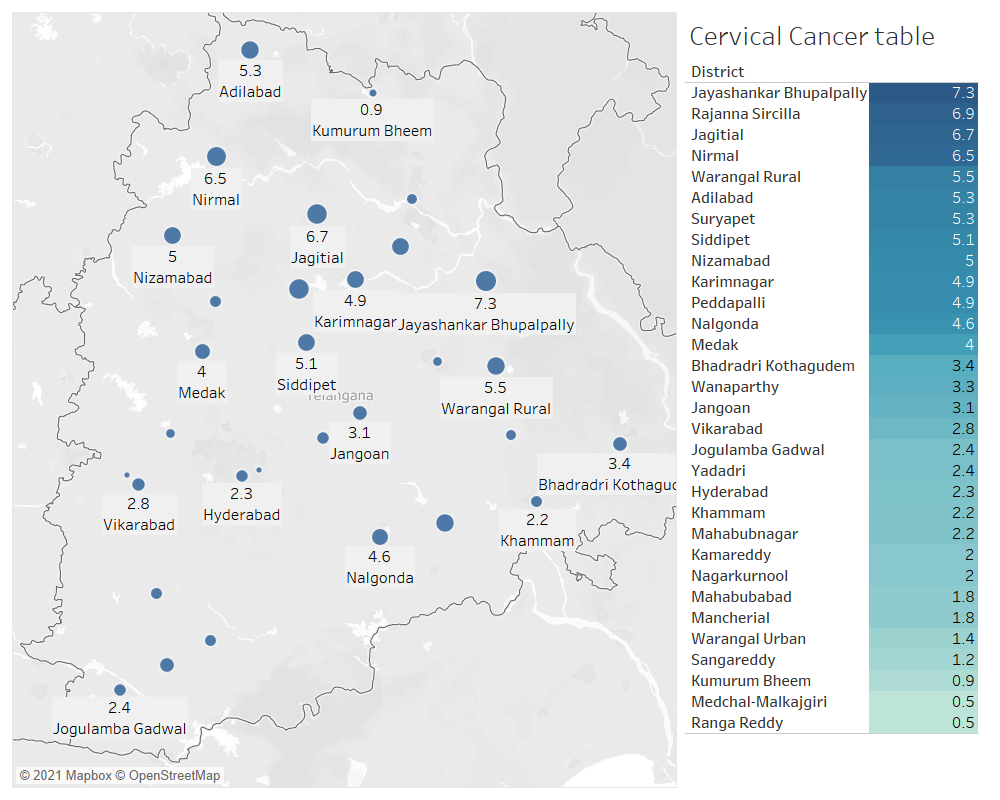}
\caption{\centering{Percentage distribution of screening tests for Cervical Cancer taken by Women in Telangana (District wise)}}
\end{figure}

\begin{figure}[!h]
\centering
\includegraphics[width=3.4in]{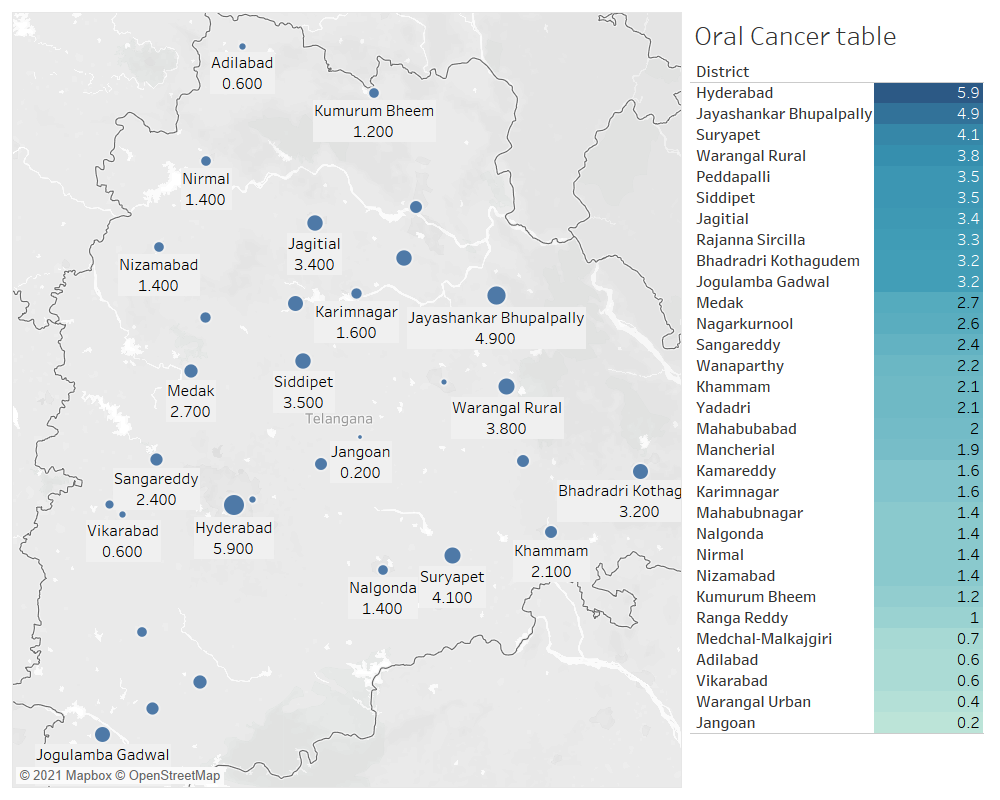}
\caption{\centering{Percentage distribution of oral cavity examination for Oral Cancer taken by Women in Telangana (District wise)}}
\end{figure}

According to the report “Profile of Cancer and Related Factors - Telangana, 2021”[2], the five leading sites of cancers in females are Breast (35.5\%), Cervix Uteri (8.7\%), Ovary (6.9\%), Corpus Uteri (5.5\%), Lung (4.1\%).
In the age group of 0 to 74 years, the cumulative risk of developing cancer in females is 1 in every 7.
(Cumulative risk is the probability that an individual will be diagnosed with cancer in the absence of any competing cause of death and assuming that the current trends prevail over time). 

On further studying the clinical extent of disease for cancer of selected anatomical sites in females, the following proportion shown in Fig.5 was found considering all the cases.

\begin{figure}[!h]
\centering
\includegraphics[width=3.4in]{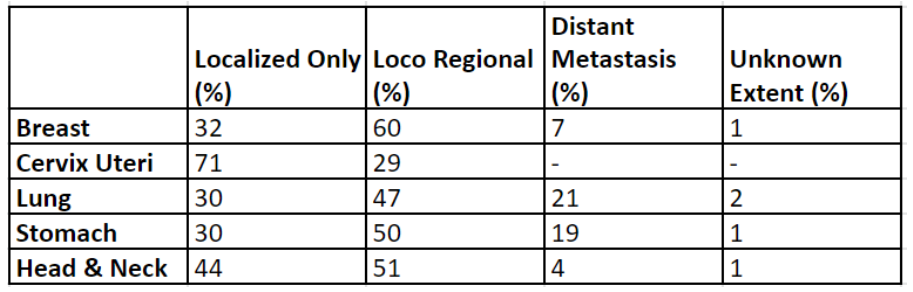}
\caption{\centering{Summary of cases with anatomical sites in females in Telangana}}
\end{figure}

A limitation of the Clinical Extent of Disease at presentation (\%) for cancers of selected anatomical sites is that it has been calculated from the HBCRs in the state which may not represent the entire state.

Projected incidence of cancer cases for the state for the years 2020 and 2025 was calculated according to gender using incidence data from the composite period of 2012-2016 was used as a reference. In the case of females, it was found to be 25434 cases in 2020 and 28708 cases in 2025.

From both of these studies/reports, we can conclude that a minuscule percentage of women undergo screening tests despite the high-risk factor. The probable reasons for this could be unawareness of the severity of existing risk, lack of accessibility to medical facilities especially in rural areas, and financial constraints. There is an immediate need for state-level and PAN India awareness programs, involving multiple stakeholders from society and the health system to improve cancer literacy in India. Thousands of lives could be saved each year if people were more aware of the signs and symptoms of cancer and people looked for help as soon as possible as treatment is usually more effective in the early stages of cancer.

Some awareness projects that already took place in Telangana are as follows:

A free cancer screening and training camp was organized at the Appolo Hospital as part of the international cancer conference going on at the hospital. As part of the camp over 60 DWCRA women were trained in procedures like self-breast examination for the early detection of cancer. According to Dr. Nalini, an oncologist, who was one of the trainers, the program aims at spreading the message across the state that cancer is a curable disease if detected early.

Dr. Charanjith Reddy Veeramalla, Managing Director of Omega Bannu Hospitals, said breast and cervical cancer were most common among Indian women and the number is increasing in recent times. However, they can be preventable if detected at an early stage through the screening tests and that there is a need to conduct more screening tests for the women for the prevention of breast and cervical cancer. The District Legal Services Authority (DLSA) of Warangal, in association with Omega Bannu Hospitals, had organized a free cancer screening test camp at the DLS building in Aug 2021 for the benefit of female judicial officers, staff members, and advocates. A total of 80 women underwent tests at the camp. Tests like Mammography for Breast Cancer(for females age 40 years above), Pap smear for Cervical Cancer(for females age 18 years above), Anaemia Screening by Blood Test and ECG, were conducted at the camp.

\section{Model Function}
The main goal of our model is the classification of patients as per their susceptibility to breast and cervical cancer. Cancer, to a certain extent, depends on environmental and demographic factors as well as genetic predisposition. Hence early screening of individuals susceptible to cancer can greatly reduce treatment needed, recovery time and provide a better chance at survival. The model developed in this project learns the relationship between these factors and the risk of developing cancer. It classifies individuals into two categories, susceptible to breast/cervical cancer and not susceptible to it. 
Susceptible individuals can then go through clinical breast exams, mammographies and ultrasounds to detect cancer early.

\section{Data} 

\subsection{Cervical Cancer}
The data set was obtained from Kaggle[3], which is gratefully acknowledged. The data was gathered at the 'Hospital Universitario de Caracas' in Caracas, Venezuela. The dataset includes 858 patients' demographics, habits, and medical records from the past. Because of privacy concerns, several patients chose not to answer certain questions (missing values). This data set describes the risk factors for cervical cancer that lead to a biopsy. This dataset is made up of 36 columns and 858 rows. It is a multivariate dataset that contains both real and integer values.  

\subsection{Breast Cancer}
The data set was obtained from BCSC Research[4] and is kindly appreciated. Originally collected by the Breast Cancer Surveillance Consortium in 1996-2002, this dataset specifies risk factors for susceptibility to breast cancer. This data record consists of 16 columns and 4,62,563 rows. It is a multivariate data set with real and integer values.  

\section{Preprocessing and features}

\subsection{Cervical Cancer}
All numerical data provided in the dataset was processed before use. There were a lot of missing values and 22 duplicate values in the dataset which were completely removed to avoid ambiguity. Certain columns with exceeding null values and no correlation were completely excluded from the dataset. All the numerical values were cast to float type using astype() in python. After cleaning the dataset, patients with no null values and duplicate values were retained in the dataset. After this processing, a dataset of 688 patients was obtained which was subsequently used for model training. The data set is split for training and validation in a ratio of 3:1 (75\% and 25\% respectively). The age, number of sexual partners, first sexual encounter, number of pregnancies, smokes(years), Hormonal Contraceptives (years), IUD (years), Dx: HPV, etc are the primary features used for modelling. The independent parameters consisted of all the columns except ‘Dx: Cancer’ from the processed dataset. The dependent parameter was the column ‘Dx: Cancer’. The independent features were fitted to avoid variance. 

\begin{figure}[!h]
\centering
\includegraphics[width=3.4in,height=2in]{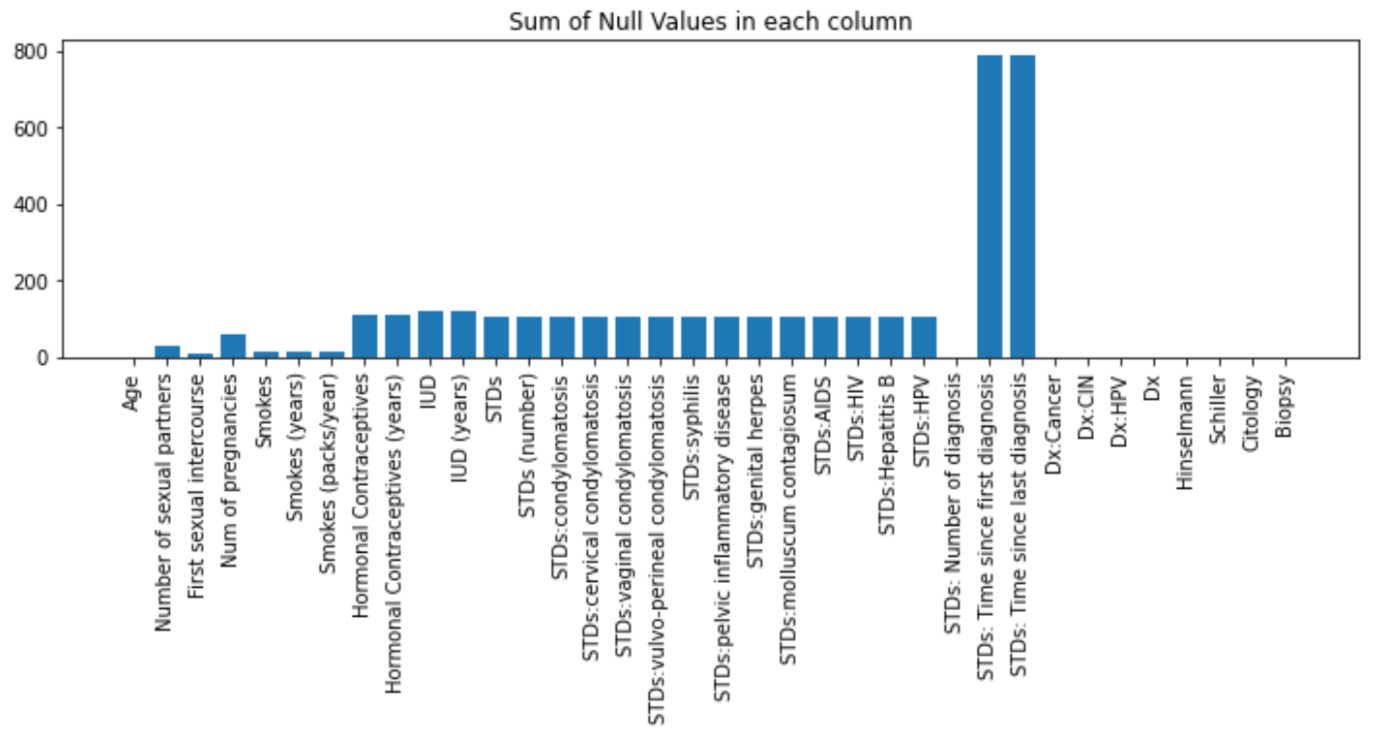}
\caption{\centering{Number of null values in every feature of the dataset}}
\end{figure}

\begin{figure}[!h]
\centering
\includegraphics[width=3.4in]{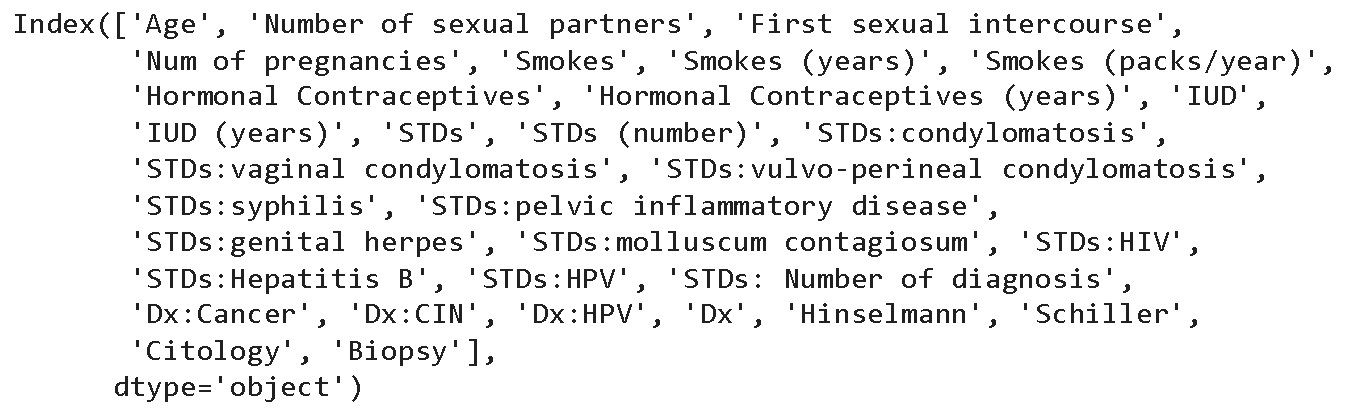}
\caption{\centering{Features used for training the ML classification model for cervical cancer}}
\end{figure}

\subsection{Breast Cancer}
The data collected as mentioned above was cleaned and processed before it was used for model fitting. The data set contained a ‘training’ column that divided the data into validation and training categories, which was removed. In addition, missing values in the data set were identified and rows with missing values were removed. Duplicate samples (14655) have also been removed. 

\begin{figure}[!h]
\centering
\includegraphics[width=3.4in,height=2in]{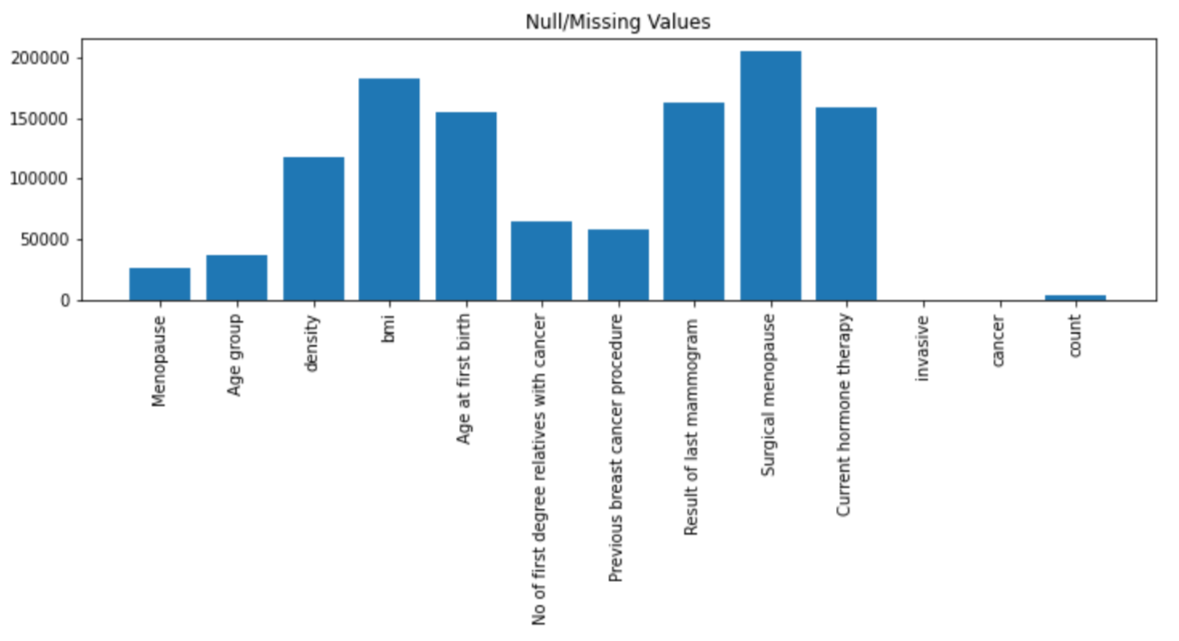}
\caption{\centering{Number of missing values in every feature of the dataset}}
\end{figure}

After cleaning, 15203 rows were left. The data was also normalised by scaling to unit variance using StandardScaler() before training. The data set was split for training and testing in a ratio of 1:1 (50\% and 50\% respectively). 

The primary features used for modelling are the age group, menopause, invasive tumour, basal metabolic index, age at first birth, number of relatives with first-degree cancer, current hormone therapy, BI-RADS density, previous breast cancer procedure, the result of the mammogram, surgical menopause and current hormone therapy. The dependent feature to be predicted is cancer. 

\section{Exploratory Data Analysis}
\subsection{Cervical Cancer}
 After loading the dataset we first look at the dimensions which are (650, 32). Looking at the information of the dataset to get insights to the data like its features, data types of the feature, etc.
 Thereafter, we preprocess and clean the data. Statistical summary of the features can be useful in inspecting the feature distribution and anomalies, if any.
\begin{itemize}
  \item The maximum value in the 'Age' column is 84, but the maximum value in the other columns is much lower, which could lead to poor model performance because the 'Age' column has more influence than the other columns. To avoid differences in influence when training the model, standardise the values of all columns.The maximum value in the 'Age' column is 84, but the maximum value in the other columns is much lower, which could lead to poor model performance because the 'Age' column has more influence than the other columns. To avoid differences in influence when training the model, standardise the values of all columns.
  \item The maximum value in the 'Num of pregnancies' column is 11, which is a very high number of pregnancies, and it is possible that this is an outlier that affects all of the other values in this column. The solution could be to delete these rows, but we will only do so if our performance is poor.
  \item The columns 'STDs:cervical condylomatosis' and 'STDs:AIDS' contain only zeros and are thus useless. Getting rid of them is the solution.
  \item Since the mean is 0.0255, the 'Dx:Cancer' column (which will be our dependent variable or variable to predict) is very unbalanced. The mean would be close to 0.5 if the class was well balanced. For a better understanding, we will depict this with a plot. Solving this problem is extremely difficult; the best solution would be to obtain more positive data to train our model, but this is not possible in our case; another solution could be to remove some negative cases to balance them with the positive cases, but this would result in a significant loss of information.
\end{itemize}

Before we can standardise our data, we need to know if we have columns that provide the same (or very similar) information, which could cause our model to perform poorly. This information can be obtained by creating a correlation matrix. We then plot a few columns against the Dx:Cancer column individually to check dependence of Cancer on each of the features.

\begin{figure}[!h]
\centering
\includegraphics[width=3.4in]{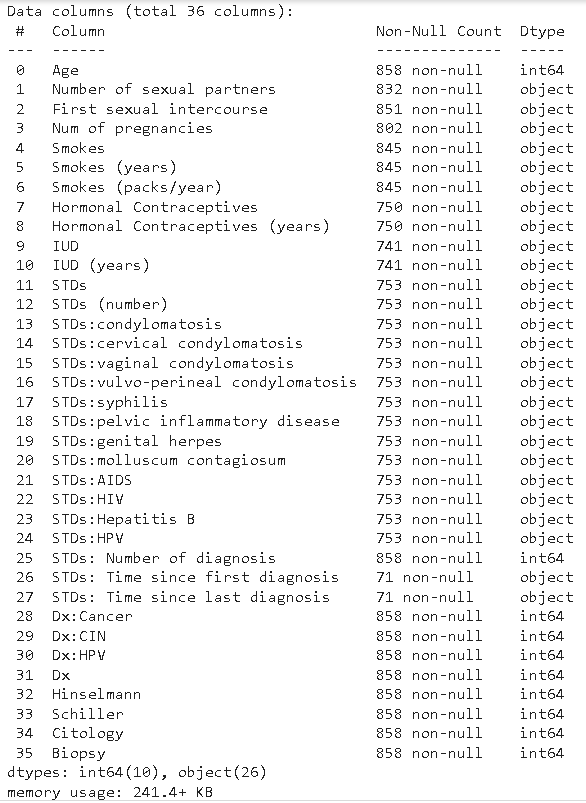}
\caption{\centering{Information of the dataset}}
\end{figure}

\begin{figure}[!h]
\centering
\includegraphics[width=3.4in,height=2.6in]{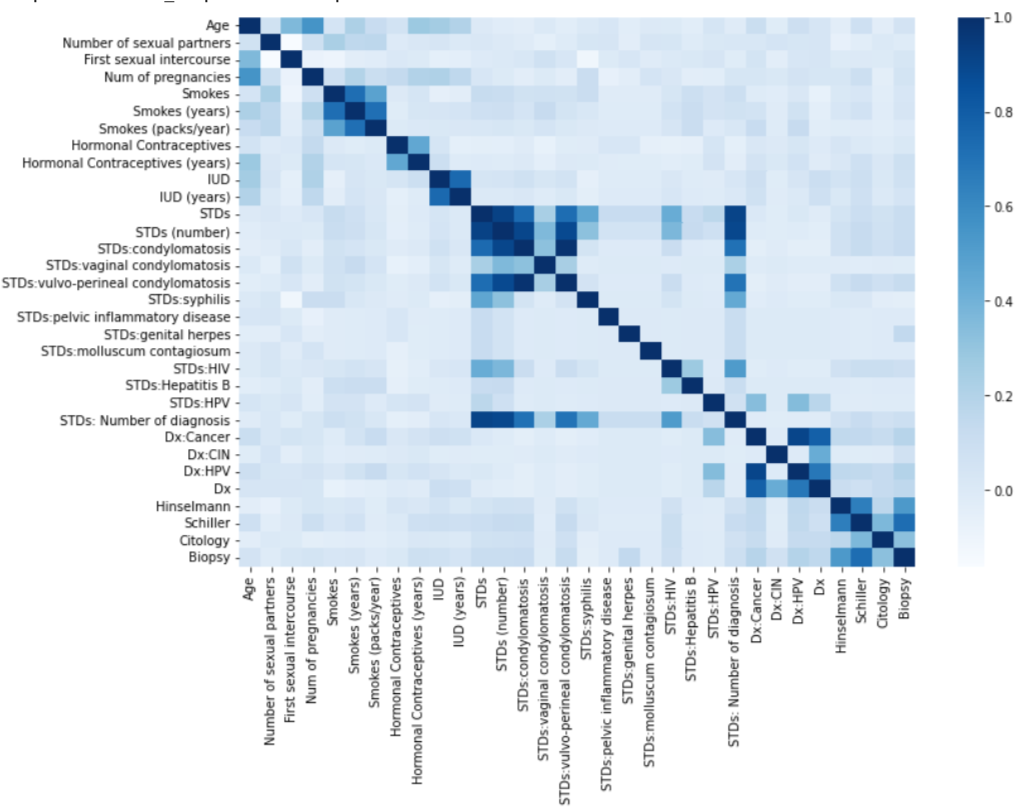}
\caption{\centering{Heatmap depicting correlation between all the dataset features}}
\end{figure}

\subsection{Breast Cancer}
After loading the dataset, we look at the dimensions. Looking at the information of the dataset to get insights into the data like its features and data types.

\begin{figure}[!h]
\centering
\includegraphics[width=3.4in]{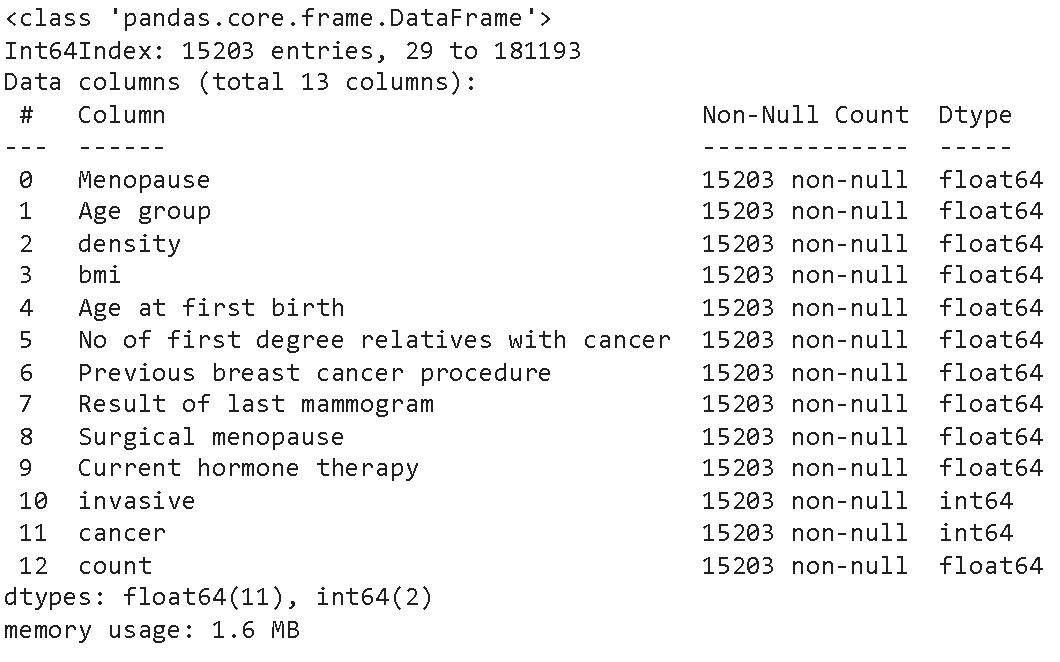}
\caption{\centering{Basic information about the features used for training}}
\end{figure}

Thereafter, we preprocess and clean the data. We drop columns that are not useful and rows with missing values. Statistical summary of the features can be useful in inspecting the feature distribution and anomalies if any.

The summary given below gives us a lot of valuable information about the data.
\begin{itemize}
  \item The maximum value in the 'count' column is 1128, significantly higher than the maximum of the other columns, which could lead to poor model performance because this column has more influence than the others. To avoid differences in influence when training the model, we standardise the values of all columns.
  \item Since the mean is 0.043, the 'cancer' column (which will be our dependent variable) is very unbalanced. If the class was balanced, the mean would have been 0.5.  For a better understanding, we will depict this with a plot. Solving this problem is extremely difficult - the best solution would be to obtain more positive data to train our model, but this is not possible in our case. Another solution could be to remove some negative cases to balance them with the positive cases, but this would result in a significant loss of information.
\end{itemize}

Before we can standardise our data, we need to know if we have columns that provide the same (or very similar) information, which could cause our model to perform poorly. This information can be obtained by creating a correlation matrix. We then plot a few columns against the cancer column individually to check the dependence of cancer on each of the features.

\begin{figure}[!h]
\centering
\includegraphics[width=3.4in,height=2.5in]{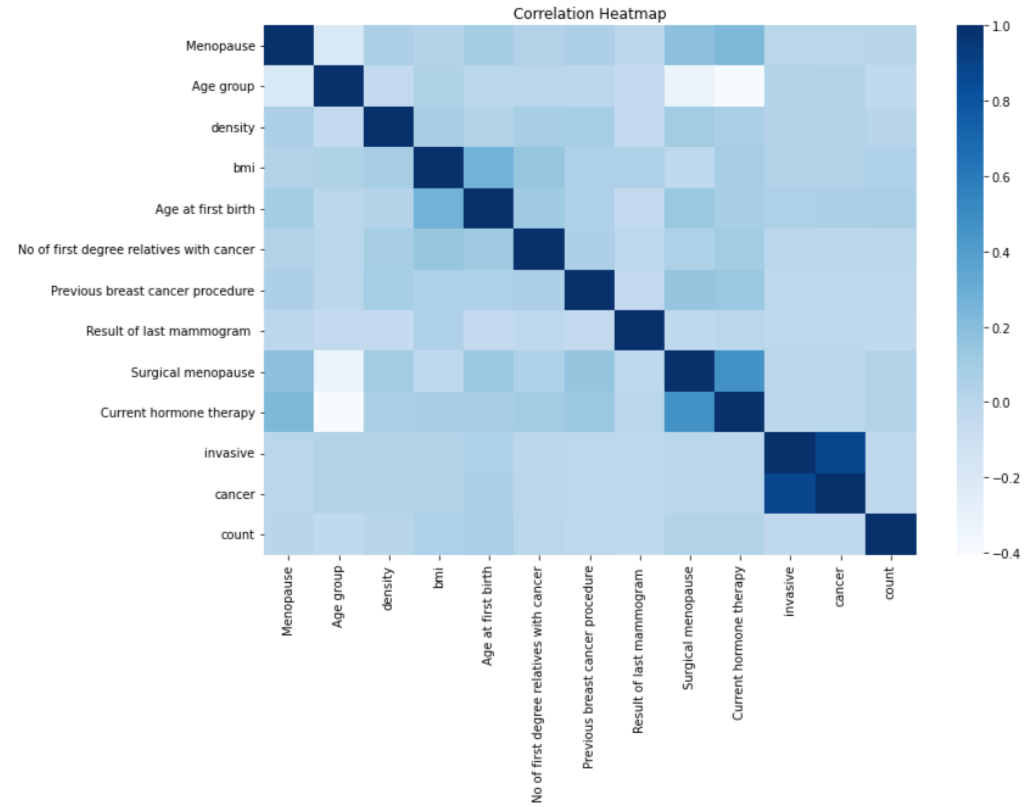}
\caption{\centering{Heatmap depicting the correlation between all  the dataset features}}
\end{figure}

\section{Classification model}

\subsection{Support Vector Classification}
A support vector machine (SVM) is a supervised machine learning model that solves two-group classification problems using classification algorithms. They have two major advantages over newer algorithms, such as neural networks, in terms of speed and performance with a limited number of samples (in the thousands). 
SVMs are particularly suitable for this task since they perform well in a high-dimensional space. SVMs are versatile in terms of the types of classification issues they are suited for since they allow the user to specify the kernel function that will act as the decision function in the model. SVMs become less successful as the number of features exceeds the number of samples, notwithstanding their effectiveness and versatility. Because this was not the case for both of our classification issues (36 labels, 688 training samples in the cervical cancer data set and 13 labels, 15203 training samples in the breast cancer data set), the group began by categorising using Support Vector Classification.

\subsection{Stochastic Gradient Descent}
Stochastic Gradient Descent (SGD) is a simple yet effective method for fitting linear classifiers and regressors to convex loss functions such as (linear) Support Vector Machines and Logistic Regression. SGD has been successfully applied to large-scale and sparse machine learning problems. Because the data is sparse, the classifiers in this module easily scale to problems with more than 105 training examples and more than 105 features. SGD, strictly speaking, is an optimization technique that does not correspond to a specific family of machine learning models. It is simply a method for training a model. SGDs are useful for this problem because they are very efficient and easy to implement.

\begin{figure}[!h]
\centering
\includegraphics[width=3.4in]{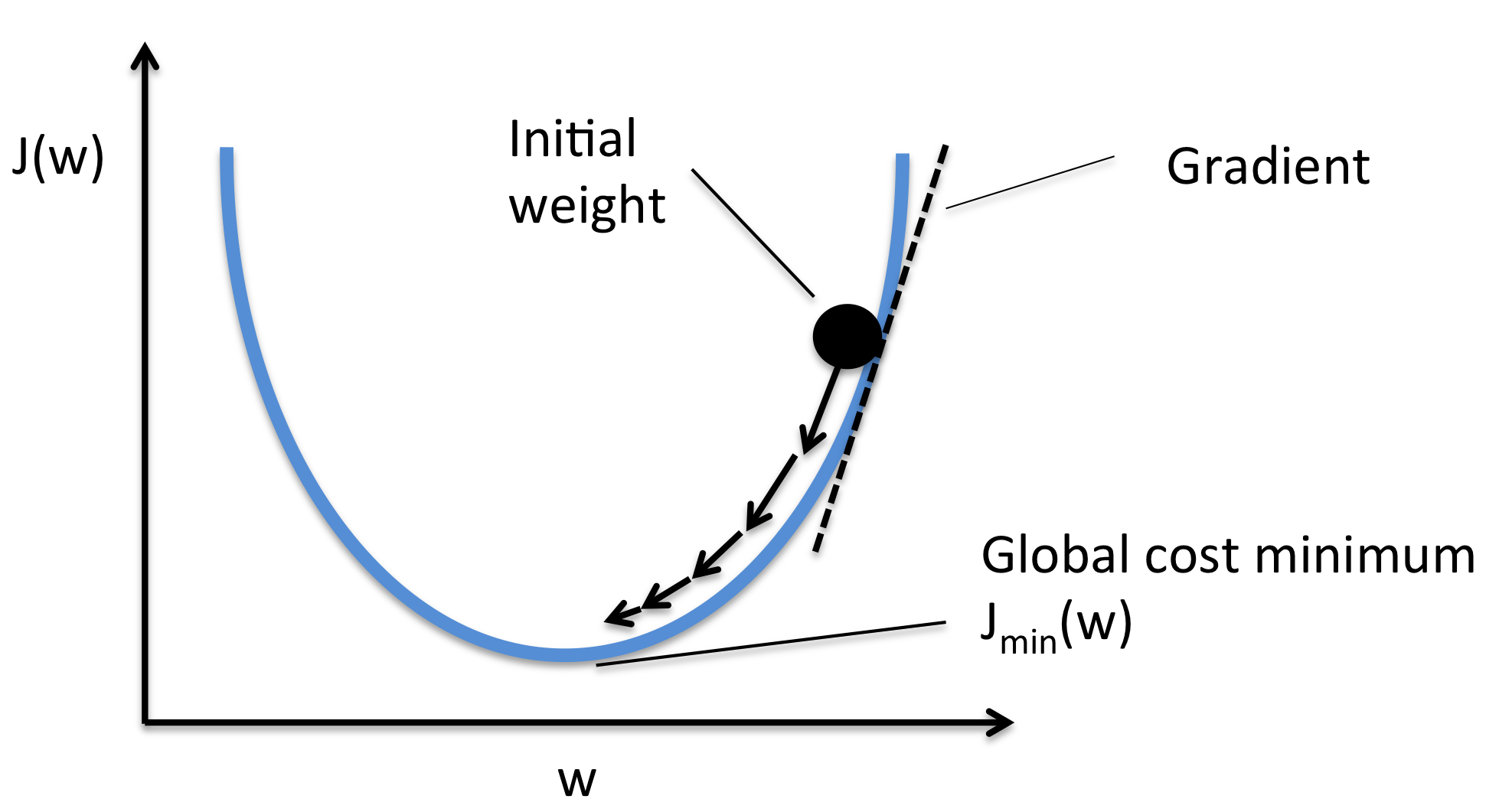}
\caption{\centering{Example of gradient descent along cost function}}
\end{figure}

SGDClassifier facilitates multi-class classification by combining multiple binary classifiers in an OVA ("one in all") scheme. A binary classifier that distinguishes between each of the k classes is learned for each of the k classes. We compute the confidence score (i.e. the signed distances to the hyperplane) for each classifier at the testing time and select the class with the highest confidence.

\subsection{Decision Tree Classifier}
The classification technique is a systematic approach to building classification models from a set of input data. Decision tree classifiers, rule-based classifiers, neural networks, support vector machines, and naive Bayesian classifiers, for example, are different techniques for solving a classification problem. Each technique adopts a learning algorithm to identify the model that best fits the relationship between the attribute set and the class label of the input data. Therefore, the main goal of the learning algorithm is to create a predictive model that accurately predicts the class names of previously unknown records.

Decision Tree Classifier is a simple and widely used classification technique. It applies a straightforward idea to solve the classification problem. Decision Tree Classifier poses a series of carefully crafted questions about the attributes of the test record. Each time it receives an answer, a follow-up question is asked until a conclusion about the class label of the record is reached.

\begin{figure}[!h]
\centering
\includegraphics[width=3.4in]{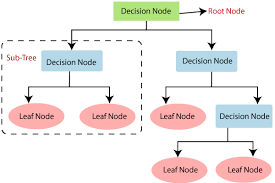}
\caption{\centering{Example of  decision tree}}
\end{figure}

\subsection{Random Forest Classifier}
Random Forest is a supervised learning algorithm. The forest built consists of multiple decision trees that are generally trained using the 'bagging' method. The general idea of the bagging method is that a combination of learning models increases the overall result. Each tree in the random forest generates a class prediction and the class with the most votes becomes the model prediction. 

Random Forest adds extra randomness to the model as the trees grow. Instead of looking for the most important feature when splitting a node, the best feature is sought out of a random subset of features. This leads to a great variety, which generally leads to a better model.

\section{Result}
We are very pleased with the final results of our classification models. All the four classification models mentioned produced the greatest train and test accuracies. The models used to train the cervical cancer dataset were stochastic gradient descent, support vector classification and decision tree classification. In the same way, the models used to train breast cancer data set were support vector classification, decision tree classification and random forest classification. In the end, we chose the best classifier for each of the datasets, the details of which are explained further.
 
\subsection{Cervical Cancer}
Out of the three classification models mentioned above, the decision tree classification algorithm produced the best results. The test data (25\%) accuracy for the same is 99.39 percent while the train data (75\%) accuracy is 100 percent. The confusion matrix and the classification report of the results produced by the decision tree algorithm is as shown in Fig. 16.

\begin{figure}[!h]
\centering
\includegraphics[width=3.4in]{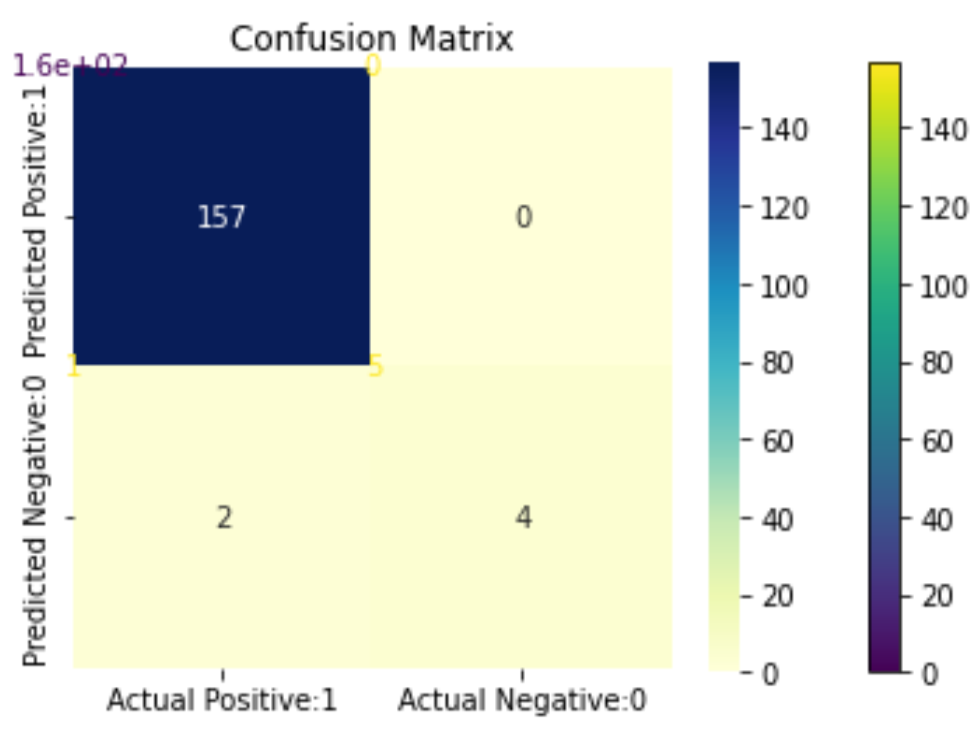}
\caption{\centering{Confusion matrix obtained after training the model using Decision Tree Classification Algorithm}}
\end{figure}

\begin{figure}[!h]
\centering
\includegraphics[width=3.4in]{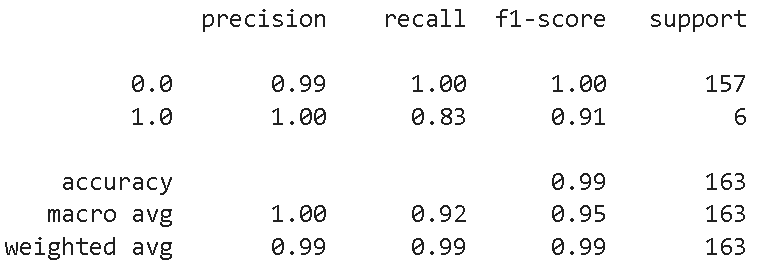}
\caption{\centering{Classification Report obtained after training the model using Decision Tree Classification Algorithm}}
\end{figure}

\subsection{Breast Cancer}
Out of the three classification models, the Support Vector Classification produced the best results. The test data (50\%) accuracy for the same is 98.89 percent while the train data (50\%) accuracy is 99.04 percent.
The confusion matrix, classification report and feature importances obtained as shown in Fig. 18.

\begin{figure}[!h]
\centering
\includegraphics[width=3.4in]{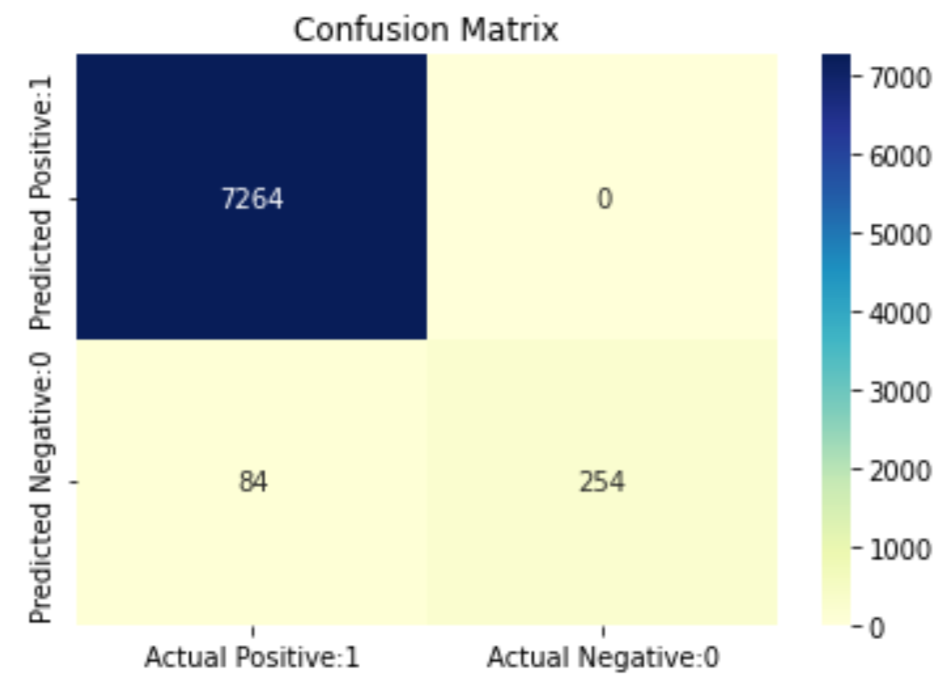}
\caption{\centering{Confusion matrix obtained after training the model}}
\end{figure}

\begin{figure}[!h]
\centering
\includegraphics[width=3.4in]{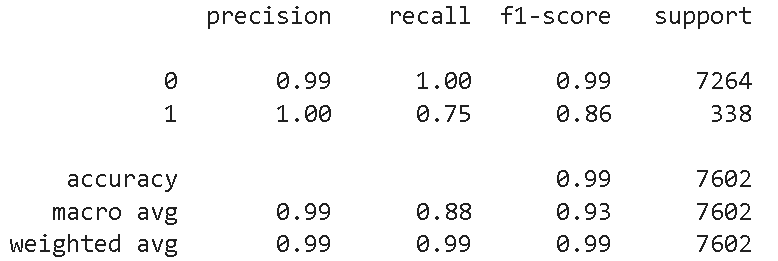}
\caption{\centering{Classification report obtained after training the model}}
\end{figure}

\section{Continued work}
The work presented in this paper devises a classification model which can be used to check one’s susceptibility for Breast and Cervical cancer. The model can be an effective tool as an open-source application available to everyone. Additionally, the application would consist of other approaches supporting our attempt towards spreading awareness and highlighting its importance.  

\subsection{Nearest Hospital/Centre suggestions}
The application will be accompanied by our system which provides suggestions for the nearest hospital or Cancer treatment centres based on the user's location or address. The system incorporates 2 APIs which help find the best suggestions. The first API from Position Stack helps find the latitude and longitude of the user and the second API from MapMyIndia helps find the nearest hospital or Cancer treatment centres. These suggestions will ensure that everyone also knows the correct source to contact in case of concern.  

\subsection{Awareness drives and Campaigns}
Cancer awareness campaigns are crucial in cancer prevention programs. The aim of these campaigns is to create cancer awareness amongst the population of Telangana. It is important to dispel the myths that people wrongly believe, inform them about the signs and symptoms, and the importance of screening for early detection. Moreover, knowledge of cancer risk factors is a determinant element in this process. 

\subsection{Integration with Health Card}
On September 27, 2021 Prime Minister Narendra Modi introduced the digital health id card which will be provided to all people. It will create a seamless online platform that will make all the health-related information portable and easily accessible to doctors. This can be used to integrate with our system for easy access. Everyone's health records will be maintained and used to identify the need for campaigns and drives based on locations and demographics using a k-means clustering algorithm as shown in Fig. 19. 

\begin{figure}[!h]
\centering
\includegraphics[width=3.4in]{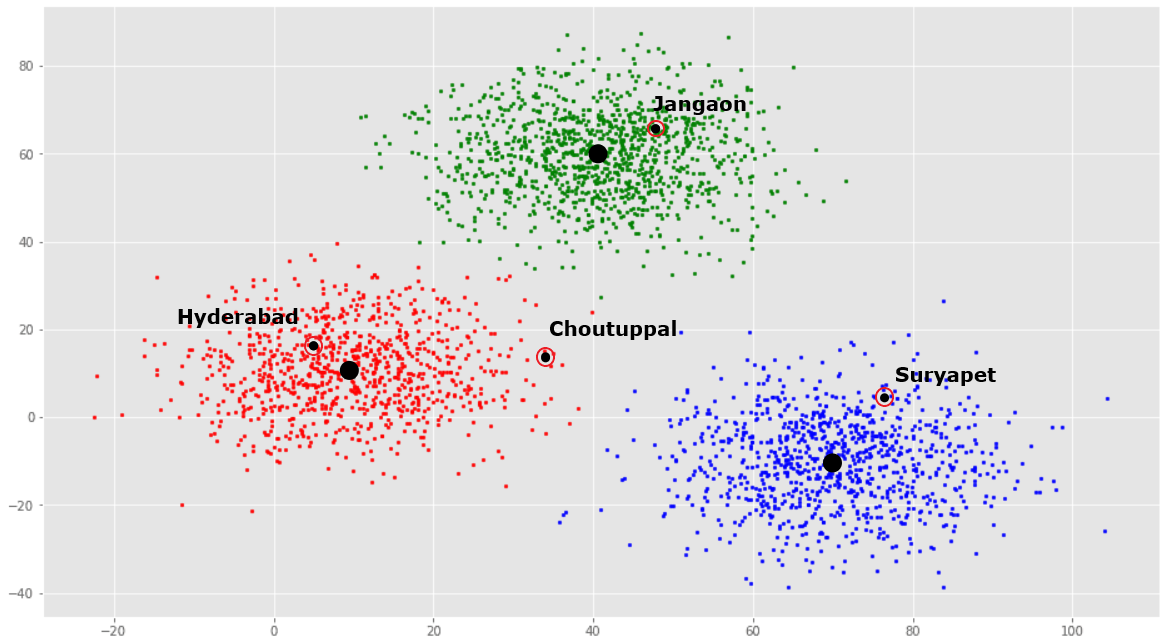}
\caption{\centering{Example of k-means clustering algorithm to determine the locations of Campaigns}}
\end{figure}

\section{CONCLUSION}
After analyzing the current statistics of cancer screening it was found that there is a critical need for Cancer Literacy in females. Awareness drives and training programs need to be frequently held for this purpose. We have devised an ML classification model to predict if a person is susceptible to breast or cervical cancer based on demographic factors. A suggestion system is then triggered which directs the user to their nearest hospital. Since this system which is integrated with susceptibility calculation and hospital suggestion is open source, people will be easily able to check their risk and get a cancer screening test if required. To take this a step further, we can integrate it with the Health Card which would make patient data more accessible (based on the privacy settings of the patients) which can help in making new policies considering real-time statistics and thereby increasing cancer literacy amongst women. Further, looking at the district-wise cancer cases and mortality, new localized schemes, awareness drives, training camps, and financial support policies can be organized which can be tailored to the level of cancer literacy and degree of urbanization in that district in Telangana.

\end{document}